\pdfoutput=1

\documentclass[sigconf]{aamas}

\usepackage{balance} %

\setcopyright{none} %
\acmConference[AAMAS '24]{Proc.\@ of the 23rd International Conference
on Autonomous Agents and Multiagent Systems (AAMAS 2024)}{May 6 -- 10, 2024}
{Auckland, New Zealand}{N.~Alechina, V.~Dignum, M.~Dastani, J.S.~Sichman (eds.)}
\copyrightyear{2024}
\acmYear{2024}
\acmDOI{}
\acmPrice{}
\acmISBN{}

\acmSubmissionID{489}

\title[Multi-Agent Reinforcement Learning for Assessing False-Data Injection Attacks on Transportation Networks]{Multi-Agent Reinforcement Learning for Assessing\\False-Data Injection Attacks on Transportation Networks}

\author{Taha Eghtesad}
\affiliation{
  \institution{Pennsylvania State University}
  \city{University Park, PA}
  \country{USA}}
\email{tahaeghtesad@psu.edu}

\author{Sirui Li}
\affiliation{
    \institution{Massachusetts Institute of Technology}
    \city{Cambridge, MA}
    \country{USA}}
\email{siruil@mit.edu}

\author{Yevgeniy Vorobeychik}
\affiliation{
    \institution{Washington University St. Louis}
    \city{St. Louis, MO}
    \country{USA}}
\email{yvorobeychik@wustl.edu}

\author{Aron Laszka}
\affiliation{
    \institution{Pennsylvania State University}
    \city{University Park, PA}
    \country{USA}}
\email{alaszka@psu.edu}

\begin{abstract}
The increasing reliance of drivers on navigation applications has made transportation networks more susceptible to data-manipulation attacks by malicious actors. Adversaries may exploit vulnerabilities in the data collection or processing of navigation services to inject false information, and to thus interfere with the drivers' route selection. Such attacks can significantly increase traffic congestions, resulting in substantial waste of time and resources, and may even disrupt essential services that rely on road networks. To assess the threat posed by such attacks, we introduce a computational framework to find worst-case data-injection attacks against transportation networks. First, we devise an adversarial model with a threat actor who can manipulate drivers by increasing the travel times that they perceive on certain roads. Then, we employ hierarchical multi-agent reinforcement learning to find an approximate optimal adversarial strategy for data manipulation. We demonstrate the applicability of our approach through simulating attacks on the Sioux Falls, ND network topology.
\end{abstract}


\usepackage{macros}

\makeatletter
\gdef\@copyrightpermission{
	\begin{minipage}{0.3\columnwidth}
		\href{https://creativecommons.org/licenses/by/4.0/}{\includegraphics[width=0.90\textwidth]{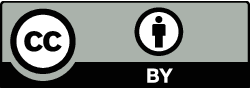}}
	\end{minipage}\hfill
	\begin{minipage}{0.7\columnwidth}
		\href{https://creativecommons.org/licenses/by/4.0/}{This work is licensed under a Creative Commons Attribution International 4.0 License.}
	\end{minipage}
	\vspace{5pt}
}
\makeatother

\begin{document}

\pagestyle{fancy}
\fancyhead{}

\maketitle

\section{Introduction}
\label{sec:intro}

In today's digitally interconnected world, drivers rely on navigation applications and online information more than before. Furthermore, the availability of social media has accelerated the spread of misinformation. A malicious actor could manipulate the drivers directly by sending malicious information through SMS messaging \cite{waniek2021traffic}, manipulating traffic signals \cite{chen2018exposing,levy2015cyber,laszka2016vulnerability,feng2018vulnerability,reilly2016creating}, or physically changing the road signs \cite{eykholt2018robust} to interfere with drivers' route selection. With the availability of social media, the drivers can further spread this misinformation to their peers to snowball the effect of manipulation. Alternatively, the adversary can inject false information into the navigation application. For example, one can place phones in a cart and pull them on the street, tampering with the navigation application to result in marking the road with heavy traffic and rerouting the drivers \cite{schoon2020google}.

Manipulating transit networks can lead to increased traffic congestion leading to devastating consequences. Modern societies heavily rely on road networks for accessing essential services such as education, healthcare, and emergency services. Moreover, road networks contribute to economic growth by enabling logistic movements of materials, goods, and products. Disruption of transportation networks can therefore lead to food insecurity, job losses, or even political disarray, such as the Fort Lee scandal \cite{wikipedia2018fort}.

Efforts have been made to measure the impact of false information injection on dynamic navigation applications \cite{lin2018data}, traffic congestion \cite{waniek2021traffic}, and navigation applications \cite{raponi2021road}. However, finding an optimal attack is in general computationally challenging~\cite{waniek2021traffic}, complicating vulnerability analysis.

The injection of false data into navigation applications is a complex task that involves several actions and decisions over time, given the dynamic nature of traffic patterns. The use of Reinforcement Learning (RL) approaches provides an effective and versatile solution to tackle such sequential decision-making challenges.

When faced with decision-making problems of any size, Reinforcement Learning (RL) can be a powerful tool~\cite{nguyen2021deep,tong2020finding}. However, the larger the problem, the more computation power is needed to train an RL model. For example, when dealing with a persistent adversary injecting false information into a city-wide transportation network of a major city, manipulating thousands of network links and millions of vehicles would make training of out-of-the-box RL strategies impractical. In such cases, a multi-agent and hierarchical RL framework becomes necessary. This framework should have local adversarial RL agents assigned to a subsection of the transportation network, observing their local information, and making local decisions. Additionally, these agents should cooperate to find the optimal sequence of false information to be injected. The agents should be coordinated with a global agent that assigns significance to each locality.

\subsection{Contributions}

Our goal is to examine the possible effects of false information injection on navigation apps. To achieve this, we created an adversarial threat model that outlines the relationships between vehicles, the navigation app, and a false information actor. Using a Markov Decision Process (MDP), we constructed this model to optimize the total travel time for all vehicles to arrive at their destinations.

We introduce a Hierarchical Multi-Agent (Deep) Reinforcement Learning (HMARL) system that consists of two levels. At the lower level, local agents observe the information of their immediate surroundings and collaborate to find the best local strategy to inject false information. The agents are constrained by an attack budget. At the higher level, a global agent coordinates the activities of the lower-level agents.

Finally, we evaluated our approach by comparing it with baseline RL algorithms and non-optimal heuristic approaches through an ablation study in Sioux Falls, ND.

\subsection{Organization}
The article is divided into several sections. In Section~\ref{sec:related}, we conduct a literature review on false information injection in navigation apps and scalable reinforcement learning techniques. In Section~\ref{sec:model}, we describe the interactions of the the adversarial threat actor and the transportation network. In Section~\ref{sec:background}, we provide the background information necessary to comprehend our approach. In Section~\ref{sec:approach}, we explain our HMARL approach. In Section~\ref{sec:experiments}, we assess the HMARL approach on a benchmark transportation network, Sioux Falls, ND. Finally, in Section~\ref{sec:discussion}, we provide a higher-level analysis of the approach and suggest future directions.

\section{Related Work}
\label{sec:related}

The literature review discusses false information injection in navigation apps and hierarchical reinforcement learning.

\subsection{Attacks on Navigation Applications}
Several studies have been conducted to assess the effects of manipulating drivers to disrupt the normal functioning of transportation systems. These attacks can be carried out through various means, such as physically tampering with traffic signals. Researchers \cite{chen2018exposing, feng2018vulnerability, laszka2016vulnerability, levy2015cyber, reilly2016creating} have all explored these approaches. For example, Laszka et~al.~\cite{laszka2016vulnerability,laszka2019detection} demonstrated that vehicles may be susceptible to altered and tampered traffic signs that can mislead them into taking the wrong path.

In a recent study, Waniek~et~al.~\cite{waniek2021traffic} surveyed approximately 3,300 participants to investigate the effects of direct manipulation of drivers through SMS notifications and invalid road signs. The results showed that fake traffic signals and SMS notifications can significantly alter predetermined travel routes. As a result, there can be up to 5,000 additional vehicles on major road networks in~Chicago.

Through an experiment conducted by Simon Weckert~\cite{schoon2020google}, it was found that navigation apps can be misled by false information. Weckert pulled a wagon with 99 smartphones while using Google Maps, which resulted in the app marking the street as heavily congested and suggesting alternative routes to drivers. This highlights the potential vulnerability of navigation apps to false information~injection.

Recent research has revealed that data manipulation can impact transportation networks and users. However, it has been difficult to determine the full extent of this impact due to computational constraints. To address this issue, a state-of-the-art hierarchical reinforcement learning algorithm is being developed, which promises to offer a feasible solution.

\subsection{Hierarchical RL Approaches}

Hierarchical Reinforcement Learning (HRL) has gained significant attention due to its applications and development. These methods have proven to be successful in tasks that require coordination between multiple agents, such as Unnamed Aerial Vehicles (UAVs) and autonomous vehicles, to complete objectives efficiently.

For instance, Yang~et~al.~\cite{yang2018hierarchical} devised a general framework for combining compound and basic tasks in robotics, such as navigation and motor functions, respectively. However, they limited the application to single-agent RL at both levels. Similarly, Chen et al. used attention networks to incorporate environmental data with steering functions of autonomous vehicles in a hierarchical RL manner so that the vehicle can safely and smoothly change lanes.

In the UAV applications, Zhang~et~al.~\cite{zhang2020hierarchical} demonstrated the success of hierarchical RL in the coordination of wireless communication and data collection of UAVs.

Although our problem is in a different domain, the fundamental ideas of these works are applicable to us since we are dealing with cooperation and coordination between adversarial agents in finding an optimal manipulation strategy in navigation applications. The study results indicate that the use of Reinforcement Learning approaches accurately modeled the effects of false information injection on navigation apps.

\section{System Model}
\label{sec:model}

In this section, we devise and formalize our threat model with respect to a transportation network environment where the adversarial agent injects false traffic information with a restricted budget with the aim of increasing the total travel time of vehicles traveling in this network.

\subsection{Environment}

The traffic model is defined by a \emph{road network} $G = (V, E)$, where $V$ is a set of nodes representing road intersections, and $E$ is a set of directed edges representing road segments between the intersections.
Each edge $e \in E$ is associated with a tuple $e = \langle t_e, b_e, c_e, p_e \rangle$, where $t_e$ is the free flow time of the edge, $c_e$ is the capacity of the edge, and $b_e$ and $p_e$ are the parameters for the edge to calculate actual edge travel time $W_e(n_e)$ given the congestion of the network, where $n_e$ is the number of vehicles currently traveling along the edge~\cite{transportationnetworks}.
Specifically, we use the following function for $W_e(n_e)$:
\begin{align}
    W_e(n_e) = t_e \times \left(1 + b_e \left(\frac{n_e}{c_e}\right)^{p_e}\right)
\end{align}

The \emph{set of vehicle trips} are given with $R$, where each trip $r \in R$ is a tuple $\langle o_r, d_r, s_r \rangle$, with $o_r \in V$ and $d_r \in V$ the origin and destination of the trip, respectively, and $s_r$ the number of vehicles traveling between the origin-destination pair $\langle o_r,d_r \rangle$. 

\subsection{State Transition}
\label{sec:state}

For each vehicle trip $r \in R$ at each time step $t \in \mathbb{N}$, \emph{vehicle location} $l_r^t \in V \cup (E \times \mathbb{N})$ represents the location of vehicle $r$ at the end of time step $t$, where the location is either a node in $V$ or a tuple consisting of an edge in $E$ and a number in $\mathbb{N}$, which represents the number of timesteps left to traverse the edge.

Each vehicle trip begins at its origin; hence $l_r^0 = o_r$. At each timestep $t \in \mathbb{N}$, for each vehicle trip $r$ that $l_r^{t-1} \in V \setminus \{ d_r \}$, i.e., the vehicle trip is at a node but has not reached its destination yet, let $\oslash^{t-1}_r = (l_r^{t-1}, e_1, v_1, e_2, v_2, \ldots, e_k, d_r)$ be a shortest path from $l_r^{t-1}$ to $d_r$ considering congested travel times $\vect{w}^t$ as edge weights.
Then $l_r^t = \langle e_1, \lfloor w_e^{t-1} \rceil \rangle$, where the travel time of edge $e$ is
\begin{align}
    w_e^t = W_e\left(\sum_{\left\{r \in R \, \middle| \, l_r^{t-1} = \langle e, \cdot \rangle \right\}} s_r \right).
\end{align}
Thus, for a trip $r$ with $l_r^{t-1} = \langle e, n \rangle$, i.e., the vehicle is traveling along an edge, if $n = 1$, that is, the vehicle is one time step from reaching the next intersection, $l_r^t = v_1$. Otherwise, $l_r^t = \langle e, n-1 \rangle$.

\subsection{Attacker Model}

At the high level, our attack model involves adversarial perturbations to \emph{observed} (rather than actual) travel times along edges~$e$, subject to a perturbation budget constraint $B \in \mathbb{R}$.
Let $a_e^t \in \mathbb{R}$ denote the adversarial perturbation to the observed travel time over the edge $e$.
The budget constraint is then modeled as $\|\vect{a}^t\|_1 \le B$, where $\vect{a}^t$ combines all perturbations over individual edges into a vector.
The observed travel time over an edge $e$ is then
\begin{align}
    \hat{w}_e^t = w_e^t + a_e^t.
\end{align}
It is this observed travel time that is then used by the vehicles to calculate their shortest paths from their current positions in the traffic network to their respective destinations.
Since we aim to develop a defense that is robust to informational assumptions about the adversary, we assume that the attacker completely observes the environment at each time step $t$, including the structure of the transit network $G$, all of the trips $R$, and the current state of each trip $l_r$. %

The attacker's goal is to maximize the total vehicle travel times, which we formalize as the following optimization problem:
\begin{align}
\label{E:attack}
    \max_{\left\{\vect{a}^1, \vect{a}^2, \ldots\right\}: \, \forall t \left( || \vect{a}^t ||_1 = B \right)} ~ \sum_{t = 0}^\infty \gamma^t \cdot \sum_{\left\{ r \in R \, | \, l_r^t \neq d_r \right\}} s_r,
\end{align}
where $\gamma \in (0, 1)$ is a temporal discount factor.

\section{Background}
\label{sec:background}

In this section, we define the terminology and background to help better understand the solution approach.

\subsection{Deep Reinforcement Learning}
\label{sec:ddpg}
Let the tuple $\langle S, A, R, T \rangle$ define a \emph{Markov Decision Process} (MDP) where $S$ denotes the state space, $A$ denotes the action space, $R(s^t, a^t)$ $\mapsto r^t \in \mathbb{R}$ is the rewarding rule for transitioning from state $s^t \in S$ by taking action $a^t \in A$ at timestep $t$, and $T(s^t, a^t, s^{t+1}) \mapsto [0, 1]$ is the probability that taking action $a^t$ in state $s^t$ will lead to state $s^{t+1} \in S$ at the next timestep.

A \emph{Deep Reinforcement Learning} (DRL) algorithm finds an approximately optimal action strategy $\pi(s^t) \mapsto a^t$ for a MDP such that it maximizes the discounted reward $\mathbb{E}\left[\sum_{\tau=0}^\infty \gamma^t \cdot r^{t + \tau} ~\middle|~ \pi\right]$.

An \emph{Action-Value} RL method learns the expected discounted value of taking an action in a state by training an approximated parameterized function $Q^\theta$ such that 
\begin{equation}Q^\theta(s^t, a^t) = \mathbb{E}\left[r^t + \gamma \max_{a'}Q^\theta(s^{t+1}, a')\right].\end{equation}
Hence, making the approximately optimal action strategy $\pi(s^t) = \argmax_{a'}Q^\theta(s^t, a')$. The training of $Q^\theta$ is based on fitting samples of \emph{experiences} that minimize the squared Bellman loss to the \emph{Temporal Difference} target
\begin{align}
    L^\theta=\mathbb{E}_{e \sim E}\left[\left(Q^\theta(s^t, a^t) - (r^t + \gamma \max_{a'}Q^\theta(s^{t+1}, a'))\right)^2\right]
\end{align}

Each experience $e = \langle s^t, r^t, a^t, s^{t+1} \rangle \in E$ is a tuple of the state $s^t$ that the agent was in at time $t$, the action it took $a^t$, the state it arrived at at the next timestep $s^{t+1}$, and the reward that it received as the result of the state transition.

With a discrete action space, one can enumerate all possible actions in the action-value function for calculating the policy ($\argmax$ $Q^\theta$) and the target value ($\max Q^\theta$) in a process called \emph{Deep-$Q$-Learning} (D$Q$L) \cite{mnih2015human}. However, with continuous action spaces, one needs to find the best action and value with a gradient search on the $Q$ function. This led to the emergence of actor-critic methods such as \emph{Deep Deterministic Policy Gradients} (DDPG) \cite{lillicrap2015continuous}. In DDPG, a separate parameterized action function $\mu^{\theta'}(s^t) \mapsto a^t$ is used that is updated based on moving the parameters in the direction of increasing the $Q$ function by gradient ascent. Specifically, the performance of policy ($J$) that needs to be maximized can be expressed as:
\begin{align}
\label{eq:ddpg_actor} 
    J^{\theta'} = \mathbb{E}_{e \sim E}\left[Q^\theta(s^t, \mu^{\theta'}(s^t))\right]
\end{align}
leading to $\pi(s^t) = \mu^{\theta'}(s^t) = \argmax_{a'}Q(s^t, a')$.

\subsection{Multi-Agent Deep Reinforcement Learning}
\label{sec:maddpg}

While both D$Q$L and DDPG perform quite well for large state spaces, they lack scalability to large action spaces where the action gradient when updating the policy diminishes, making the policy function impossible to train. Another approach to scalability is to split the environment into $K$ disjoint components and assign one DRL agent to find an approximate optimal policy for the particular component, given the observation from the component, while the agents receive separate rewards.

One such Multi-Agent Reinforcement Learning algorithm is the \emph{Multi-Agent Deep Deterministic Policy Gradient} MADDPG~\cite{lowe2017multi} that follows a \emph{centralized training decentralized execution} model where the training of the component's $Q$ function requires access to global state information while execution of the policy function $\mu$ is done by only relying on local observations pertaining to the component. In MADDPG, each agent $i$ has a $Q^{\theta_i}(\vect{s^t}, o_i^t, a_1^t, a_2^t, \cdots, a_k^t) \forall_{k \in K}$ where $\vect{s}$ that is a joint representation of the state of the system, $o_i^t$ is the observation of agent $i$ from its component, and $a_k^t$ is the action taken by each agent $k$. $Q^{\theta_i}$ predicts the estimated discounted rewards for agent $i$ and can be updated by reducing the Bellman loss to the temporal difference target. The policy of agent $i$ is a function approximator $\mu$ parameterized by $\theta'_i$ that can be trained by assuming a similar performance function $J^{\theta'_i}$ to DDPG (Equation~\ref{eq:ddpg_actor}):
\begin{align}
    J^{\theta'_i} = \mathbb{E}_{e \sim E}\left[Q^{\theta_i}\left(\vect{s^t},o_i^t, a_1^t, a_2^t, \cdots, a_k^t, \mu^{\theta'_i}(o_i^t)\right)\right] \forall_{k \neq i}
\end{align}

\section{Hierarchical Multi-Agent Reinforcement Learning}
\label{sec:approach}

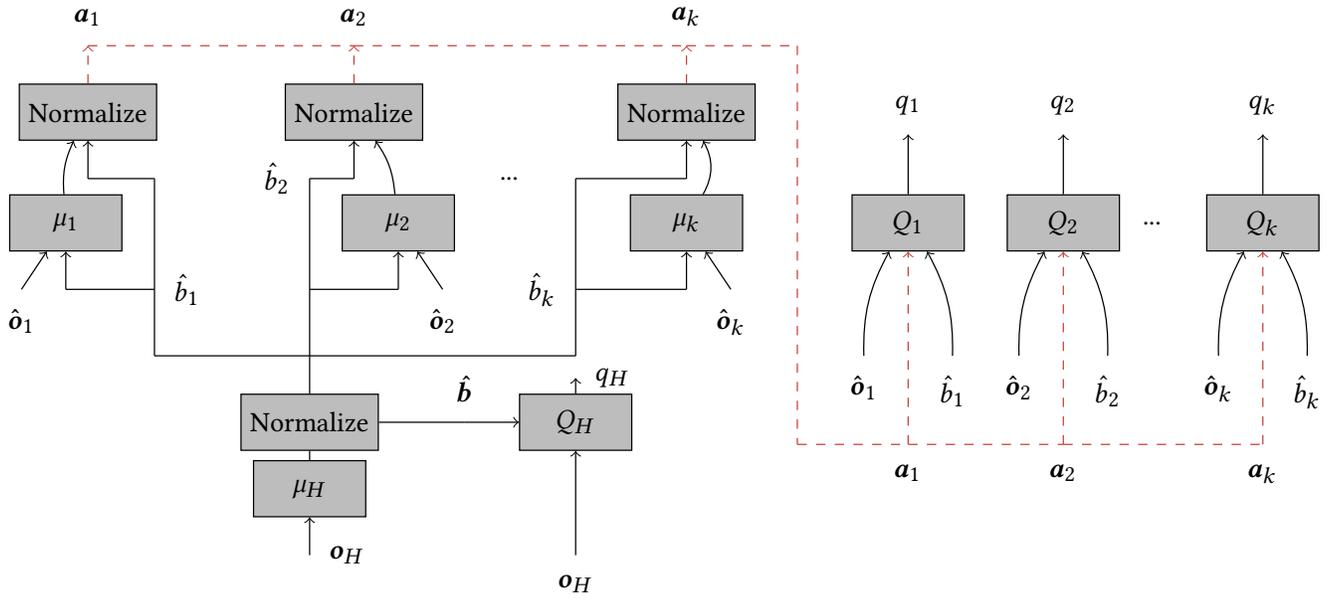
\begin{figure*}[t]
    \centering
    \resizebox{\linewidth}{!}{\begin{tikzpicture}
	\begin{pgfonlayer}{nodelayer}
            \node [style=box] (56) at (-5.5, 3.25) {Normalize};
		\node [style=box] (0) at (-5.5, 2.5) {$\mu_H$};
		\node [style=box] (1) at (-8.25, 5.5) {$\mu_1$};
		\node [style=box] (2) at (-4.5, 5.5) {$\mu_2$};
		\node [style=box] (3) at (-1.25, 5.5) {$\mu_k$};
		\node [style=box] (4) at (-8, 6.75) {Normalize};
		\node [style=box] (5) at (-5, 6.75) {Normalize};
		\node [style=box] (6) at (-1.25, 6.75) {Normalize};
		\node [style=none] (7) at (-7.25, 4) {};
		\node [style=none] (8) at (-5.5, 4) {};
		\node [style=none] (9) at (-2.5, 4) {};
		\node [style=none] (10) at (-2.5, 4.75) {};
		\node [style=none] (11) at (-2.5, 6) {};
		\node [style=none, label={left:$\hat{b}_k$}] (12) at (-2.5, 4.75) {};
		\node [style=none] (13) at (-1.25, 4.75) {};
		\node [style=none] (14) at (-1.25, 6) {};
		\node [style=none] (15) at (-5.5, 4.75) {};
		\node [style=none, label={left:$\hat{b}_2$}] (16) at (-5.5, 6) {};
		\node [style=none] (17) at (-5, 6) {};
		\node [style=none] (18) at (-4.5, 4.75) {};
		\node [style=none] (19) at (-8.25, 4.75) {};
		\node [style=none, label={right:$\hat{b}_1$}] (21) at (-7.25, 4.75) {};
		\node [style=none] (22) at (-7.25, 6) {};
		\node [style=none] (23) at (-3.25, 6) {...};
		\node [style=none, label={above:$\vect{a}_1$}] (24) at (-8, 7.5) {};
		\node [style=none, label={above:$\vect{a}_2$}] (25) at (-5, 7.5) {};
		\node [style=none, label={above:$\vect{a}_k$}] (26) at (-1.25, 7.5) {};
		\node [style=none] (27) at (0, 7.5) {};
		\node [style=none] (28) at (0, 3) {};
		\node [style=none, label={below:$\vect{a}_1$}] (29) at (1.25, 3) {};
		\node [style=none, label={below:$\vect{a}_2$}] (30) at (3, 3) {};
		\node [style=none, label={below:$\vect{a}_k$}] (31) at (5.25, 3) {};
		\node [style=box] (32) at (1.25, 5.5) {$Q_1$};
		\node [style=box] (33) at (3, 5.5) {$Q_2$};
		\node [style=box] (34) at (5.25, 5.5) {$Q_k$};
		\node [style=none, label={below:$\hat{b}_1$}] (35) at (1.75, 4) {};
		\node [style=none, label={below:$\hat{b}_2$}] (36) at (3.5, 4) {};
		\node [style=none, label={below:$\hat{b}_k$}] (37) at (5.75, 4) {};
		\node [style=none, label={below:$\vect{\hat{o}}_k$}] (38) at (4.75, 4) {};
		\node [style=none, label={below:$\vect{\hat{o}}_2$}] (39) at (2.5, 4) {};
		\node [style=none, label={below:$\vect{\hat{o}}_1$}] (40) at (0.75, 4) {};
		\node [style=box] (41) at (-2.5, 3.25) {$Q_H$};
		\node [style=none] (42) at (-8, 6) {};
		\node [style=none, label={below:$\vect{\hat{o}}_2$}] (45) at (-4, 4.75) {};
		\node [style=none, label={below:$\vect{\hat{o}}_1$}] (46) at (-8.75, 4.75) {};
		\node [style=none, label={below:$\vect{\hat{o}}_k$}] (47) at (-0.75, 4.75) {};
		\node [style=none, label={right:$\vect{o}_H$}] (48) at (-5.5, 1.75) {};
		\node [style=none, label={above:$\vect{\hat{b}}$}] (49) at (-3.75, 3.25) {};
		\node [style=none] (50) at (4, 5.5) {...};
		\node [style=none, label={below:$\vect{o}_H$}] (51) at (-2.5, 1.75) {};
		\node [style=none, label={right:$q_H$}] (52) at (-2.5, 3.75) {};
		\node [style=none, label={above:$q_1$}] (53) at (1.25, 6.5) {};
		\node [style=none, label={above:$q_2$}] (54) at (3, 6.5) {};
		\node [style=none, label={above:$q_k$}] (55) at (5.25, 6.5) {};
	\end{pgfonlayer}
	\begin{pgfonlayer}{edgelayer}
		\draw (0) to (56);
            \draw (56) to (8.center);
		\draw (8.center) to (15.center);
		\draw (15.center) to (18.center);
		\draw (8.center) to (7.center);
		\draw (7.center) to (21.center);
		\draw (21.center) to (19.center);
		\draw [in=270, out=90] (21.center) to (22.center);
		\draw (16.center) to (15.center);
		\draw (17.center) to (16.center);
		\draw (8.center) to (9.center);
		\draw (9.center) to (12.center);
		\draw (12.center) to (13.center);
		\draw (12.center) to (11.center);
		\draw (11.center) to (14.center);
		\draw [style=arrow] (14.center) to (6);
		\draw [style=arrow] (13.center) to (3);
		\draw [style=arrow] (18.center) to (2);
		\draw [style=arrow] (17.center) to (5);
		\draw [style=arrow] (19.center) to (1);
		\draw [style=arrow] (42.center) to (4);
		\draw [style=dasharrow] (4) to (24.center);
		\draw [style=dasharrow] (5) to (25.center);
		\draw [style=dasharrow] (6) to (26.center);
		\draw [style=dashline] (24.center) to (25.center);
		\draw [style=dashline] (25.center) to (26.center);
		\draw [style=dashline] (26.center) to (27.center);
		\draw [style=dashline] (27.center) to (28.center);
		\draw [style=dashline] (28.center) to (29.center);
		\draw [style=dashline] (29.center) to (30.center);
		\draw [style=dashline] (30.center) to (31.center);
		\draw [style=dasharrow] (29.center) to (32);
		\draw [style=dasharrow] (30.center) to (33);
		\draw [style=dasharrow] (31.center) to (34);
		\draw [style=arrow, bend left=15] (39.center) to (33);
		\draw [style=arrow, bend right=15] (36.center) to (33);
		\draw [style=arrow, bend left=15] (38.center) to (34);
		\draw [style=arrow, bend right=15] (37.center) to (34);
		\draw [style=arrow] (45.center) to (2);
		\draw [style=arrow] (46.center) to (1);
		\draw [style=arrow] (47.center) to (3);
		\draw [style=arrow, bend left=15] (1) to (4);
		\draw [style=arrow] (48.center) to (0);
		\draw [style=line] (56) to (49.center);
		\draw [style=arrow] (49.center) to (41);
		\draw [style=arrow] (51.center) to (41);
		\draw [style=arrow] (41) to (52.center);
		\draw [style=arrow] (33) to (54.center);
		\draw [style=arrow] (32) to (53.center);
		\draw [style=arrow] (34) to (55.center);
		\draw [style=arrow, bend right=15] (2) to (5);
		\draw [style=line] (22.center) to (42.center);
		\draw [style=arrow, bend right=15] (35.center) to (32);
		\draw [style=arrow, bend left=15] (40.center) to (32);
  		\draw [style=arrow, bend right] (3) to (6);
	\end{pgfonlayer}
\end{tikzpicture}}
    \caption{Hierarchical Multi-Agent Deep Reinforcement Learning Architecture. $\mu_H$ and $Q_H$ are the high-level agent's actor and critic function approximators, respectively. $\mu_k$ and $Q_k$ are the actor and critic function approximators of low-level agent $k$, respectively. $\vect{a} = \langle \vect{a_1} \times \hat{b}_1, \vect{a_2} \times \hat{b}_2, \cdots \vect{a_k} \times \hat{b}_k\rangle$ is the perturbations of all edges of the transit graph $G$ where $a_k$ is the perturbations of edges in component $k$. $\vect{o_k}$ and $\hat{b}_k$ are the observation of the $k$-th agent from its component and the proportion of budget allocated to it, respectively. The \emph{Normalize} layer can be constructed using the \emph{Softmax} function or the 1-norm normalization of ReLU-activated actor outputs.}
    \label{fig:architecture}
\end{figure*}

At each timestep of the game, the adversary needs to find the approximately optimal perturbations to all the edges in a city network~$G$.

The action space for the low-level agent is $|E|$-dimensional. Given a moderate-sized city such as Anaheim, CA or Chicago, IL, that has 914 and 2950 road links, respectively \cite{transportationnetworks}, it is infeasible for a Single-Agent RL algorithm to learn the optimal budget allocation~strategy.

This requires that the transit network be broken down into components. Then, an RL agent will be responsible for the edges in the component, observing the information pertaining to the component and only finding the optimal perturbations for that~component.

Approaches such as MADDPG will fail in this scenario as the agents will compete over the budget, making the MDP difficult to learn. This makes the need to devise a two-level hierarchical multi-agent reinforcement learning algorithm where the purpose of the high-level agent is to allocate the budget to the components, eliminating the competition over budget, and the purpose of the low-level agent, which itself is comprised of component agents, is to further allocate the perturbation budget between the edges in their component constrained to the allocated budget to the component by the high-level agent.

\subsection{K-Means Node Clustering}

\begin{figure*}[t]
    \centering
    \includegraphics[width=\textwidth]{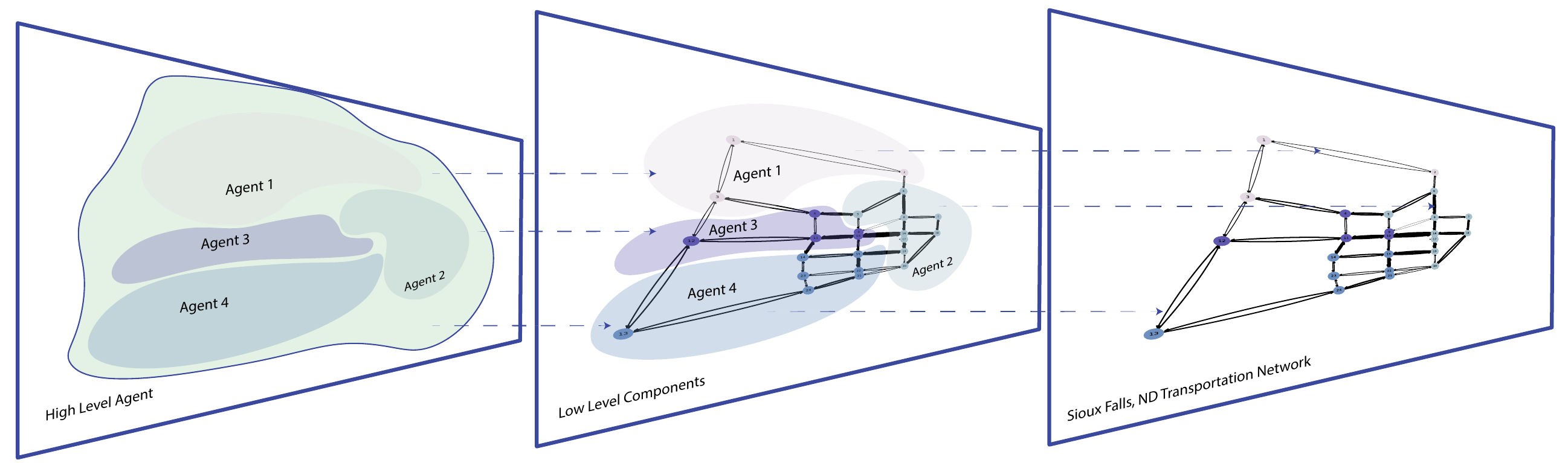}
    \caption{Decomposition of Sioux Falls, ND transportation network into four components, where one low-level agent is responsible for adding perturbation to edges in each component, and one high-level agent is responsible for allocating budget $B$ to each low-level agent. Edge width represents the density of vehicles moving over the edge without any attacker perturbation added.}
    \label{fig:decomposition}
\end{figure*}

First, these components can be formed by applying a K-Means clustering algorithm, assuming the distance between two nodes is the shortest path distance given edge weights $w_e = t_e$. Then, each edge $e=uv$ is assigned to the component of its source node~$u$. Algorithm~\ref{alg:kmeans} shows a pseudocode for the $K$-means clustering algorithm. Figure~\ref{fig:decomposition} shows the decomposition of the Sioux Falls, ND transportation network with K-Means clustering into four components.

\begin{algorithm}
\caption{$K$-Means Graph Clustering}\label{alg:kmeans}
\begin{algorithmic}
\Require A road network graph $G=(V, E)$
\State Calculate all-pairs shortest path distance $d_{u,v}:\forall_{u,v \in V}$.
\State Select $|K|$ initial nodes as component centers arbitrarily and call them $c_k \in K$.
\For{\emph{n\_iterations}}
    \State $c_u \gets \argmin_v d_{u,v}:\forall_{v \in C}$.
    \State $c_k \gets \argmin_{c_{k'}} \argmax_{u} d_{u,c_{k'}}$ such that $c_u = c_k$
\EndFor
\State $c_e = \langle uv \rangle \gets c_u \forall e \in E$ 

\Return $c_e$ for all ${e \in E}$ the centroid node for all edges.    
\end{algorithmic}
\end{algorithm}

\subsection{High and Low Level DRL Agents}

We assume that the adversarial agent has access to all features of the transportation network $G$ and all rider information $l_r^t$ at all time steps. Thus, it can summarize the information into features that can be used to train the high and low-level agents. Figure~\ref{fig:architecture} summarizes our HMARL architecture.

\subsubsection{Low-Level Multi-Agent MADDPG}

When graph $G$ is broken down into $|K|$ components, the agent supervising component $k=\hat{G}(\hat{V}_k, \hat{E}_k) \subset G(V, E)$ observes a feature vector of $\vect{\hat{o}}_k^t = \langle\langle s_e, n_e, \hat{s}_e, m_e, \tilde{s}_e \rangle : \forall_{e \in \hat{E}_k}\rangle$, where $s_e = \sum\{s_r | l_r^t \in V \wedge e \in \oslash_r^t : \forall_{r \in R}\}$ is the number of vehicles that are currently at a node with an unperturbed shortest path to the destination passing through $e$, $\hat{s}_e = \sum\{s_r | l_r^t \in V \wedge e = \oslash_r^t(e_1) :\forall_{r \in R}\}$ is the number of such vehicles that will immediately take $e$, $m_e = \sum\{s_r \cdot n | l_r^t =\langle e', n\rangle \wedge e' = e : \forall_{r \in R}\}$ is the sum of required timesteps for vehicles traveling $e$ to arrive at its endpoint, and $\hat{s}_e = \sum\{s_r | e \in \oslash_r^{t-1} : \forall_{r \in R}\}$ is the number of all vehicles taking $e$ as their shortest path at some timestep assuming the perceived travel times to remain unchanged. The agent then outputs a vector of perturbations $\vect{a}_k^t = \langle a_e^t | e \in \hat{E}_k\rangle$ to perturb all the components' edges. This agent would receive a reward $r_k^t = \sum \{s_r | l_r^t \in \hat{G}(\hat{V}_k \hat{E}_k) : \forall_{r \in R}\}$ as the number of vehicles in its component.

As the low-level agents participate in a cooperative setting with our hierarchical approach, they do not need to see other agents' actions to train their critics. The $Q$ function for each agent can be constructed with $Q^k(\vect{\hat{o}}_k^t, \vect{\hat{b}}^t_k, \vect{a}_k^t)$ with a \emph{Multi-Layer Perceptron}~(MLP) such that its output is activated with a \emph{Rectified Linear Unit}~(ReLU) as the reward for each component is non-negative, i.e., the number of vehicles in the component. The agent's action function $\mu_k(\vect{\hat{o}}_k^t, \vect{\hat{b}}^t_k) \mapsto \vect{a}_k^t$ can be constructed using an MLP. As the output of the actor function of $k$-th low-level agent needs to sum to \textbf{$\hat{b}_k$} to satisfy the budget and allocation constraint, it needs a normalizing function that can be either a Softmax function or 1-norm normalizer. The final perturbations can then be drawn by multiplying budget of the component to its action output $\vect{a} = \langle \vect{a_1} \times \hat{b}_1, \vect{a_2} \times \hat{b}_2, \cdots \vect{a_k} \times \hat{b}_k\rangle$. The training of $\mu_k$ and $Q_k$ functions can be performed according to the MADDPG algorithm (Section~\ref{sec:maddpg}).

\subsubsection{High-Level DDPG Agent}

The high-level agent $H$ observes an aggregated observation of the components at time $t$, specifically the number of vehicles in the component and number of vehicles that are making a decision in that component $\vect{o}^t_H = \langle \langle \sum_e^{\hat{E}_k} n_e, \sum_e^{\hat{E}_k} \hat{s}_e \rangle : \forall_{k \in K}\rangle$, and outputs $\vect{\hat{b}} \in [0, B]^{|K|}$ such that $\|\vect{\hat{b}}^t\|_1 = B$ the portion of the budget allocated to each component. The high-level agent is rewarded by the total number of the vehicles in the network $r^t_H = \sum_{\left\{ r \in R |l_r^t \neq d_r \right\}} s_r$. 

Similar to the low-level agent actors, the output of the high-level actor is normalized with either Softmax or 1-norm and then multiplied by the total budget $B$ to allocate each budget to the component. The training of the high level $\mu_H$ and $Q_H$ can be performed according to the DDPG algorithm (Section~\ref{sec:ddpg}).

\section{Experiments}
\label{sec:experiments}

We simulated the framework using benchmark data from \cite{transportationnetworks} and evaluated the effectiveness of HMARL for finding an optimal strategy for false information injection on the Sioux Falls, ND testbed.

\subsection{Experimental Setup}

To make the environment non-deterministic, we randomly increased or decreased $r_s$ by 5\% at each training episode's beginning. We simulate the environment by following a vehicle-based simulation based on the state transition rules of Section~\ref{sec:state}.

\subsubsection{Hardware and Software Stack} The experiments, including the neural network operations, are done on an Apple  MacBook Pro 2021 with an M1 Pro SoC with eight processing cores and 16GB of RAM. None of the experiments, including the neural operations, have been done on the Metal Performance Shaders. The simulation of the environment has been implemented using Python. For neural network operations, we used PyTorch~\cite{paszke2019pytorch}. We used NumPy~\cite{harris2020array} as our scientific computing library. The source code is available at \cite{source}.

\subsubsection{Seeds and Hyperparameters}
To make sure that the results presented in this article are reproducible, we initialized the random seeds of Numpy, PyTorch, and Python to zero. The hyperparameters used for the simulation and the training of high and low-level agents are presented in Table~\ref{tab:hyperparameters} and the neural network architectures are presented in Table~\ref{tab:nn_arch}.

\subsection{Heuristics}
\label{sec:heuristics}

We used a \emph{Greedy} heuristic as our baseline strategy. In the greedy approach, the adversarial agent counts the number of vehicles $s_e$ passing through each edge $e$ as their unperturbed shortest path to their destination. Then its applied perturbation will be $$\vect{a} = \frac{\langle s_e :\;\forall e \in E\rangle}{\sum_{e \in E} s_e} \times B.$$

When running the ablation study (see Figure~\ref{fig:siouxfalls}) and testing the high-level and low-level agents separately, we replaced the high-level with a \emph{proportional allocation}, meaning that each component agent gets a proportion of the budget relative to the number of vehicles making a decision in that component. Further, the low-level agent can be replaced with a local greedy that perturbs the edges in its component relative to the number of vehicles passing through the edges: $$a_k = \frac{\langle s_e :\;\forall e \in E_k\rangle}{\sum_{e \in E} s_e} \times \hat{b}_k.$$

\subsection{Numerical Results}

\begin{table}[t]
    \centering
    \caption{List of Hyperparameters}
    \label{tab:hyperparameters}
        \begin{tabular}{|l|c|}
            \hline
            Hyperparameter & Value \\
            \hline
            \rowcolor{Gray} \multicolumn{2}{|c|}{Environment} \\
            Training Horizon & 400 \\
            Evaluation Horizon & 50 \\
            $|K|$ Number of Components & 4 \\
            Total Training Steps & 200,000 \\
            Randomizing Factor of Number of Vehicles & 0.05 \\
            \hline
            \rowcolor{Gray} \multicolumn{2}{|c|}{Common} \\
            \hline
            $\tau_H$, $\tau_k$ target network transfer rate & 0.001 \\
            Training Batch Size & 64 \\
            Experience Replay Buffer Size & 50,000 \\
            \hline
            \rowcolor{Gray} \multicolumn{2}{|c|}{Stand Alone High Level} \\
            \hline
            $\mu_k$ Learning Rate & 0.00005\\
            $Q_k$ Learning Rate & 0.01 \\
            $\gamma_H$ & 0.99 \\
            $\tau_H$ & 0.001 \\
            Noise Decay Steps & 10,000 \\
            \hline
            \rowcolor{Gray} \multicolumn{2}{|c|}{Stand Alone Low Level} \\
            \hline
            $\mu_k$ Learning Rate & 0.00005\\
            $Q_k$ Learning Rate & 0.01 \\
            $\gamma_k$ & 0.99 \\
            Noise Decay Steps & 30,000 \\
            \hline
            \rowcolor{Gray} \multicolumn{2}{|c|}{Hierarchical} \\
            \hline
            \rowcolor{LightGray} \multicolumn{2}{|c|}{Low-Level} \\
            $\mu_k$ Learning Rate & 0.00005\\
            $Q_k$ Learning Rate & 0.01 \\
            $\gamma_k$ & 0.9 \\
            Noise Decay Steps & 10,000 \\
            \rowcolor{LightGray} \multicolumn{2}{|c|}{High-Level} \\
            $\mu_H$ Learning Rate & 0.00001\\
            $Q_H$ Learning Rate & 0.001 \\
            $\gamma_H$ & 0.99 \\
            $\tau_H$ & 0.001 \\
            Noise Decay Steps & 30,000 \\
            \hline
            \rowcolor{Gray} \multicolumn{2}{|c|}{Standalone DDPG} \\
            \hline
            $\mu$ Learning Rate & 0.00001\\
            $Q$ Learning Rate & 0.001 \\
            $\gamma$ & 0.99 \\
            Noise Decay Steps & 30,000\\
            \hline
        \end{tabular}
\end{table}

\begin{figure*}
\centering

    \begin{subfigure}[b]{0.25\textwidth}
        \centering
        \resizebox{\linewidth}{!}{\begin{tikzpicture}

\definecolor{darkgray176}{RGB}{176,176,176}
\definecolor{darkorange25512714}{RGB}{255,127,14}

\begin{axis}[
tick align=outside,
tick pos=left,
x grid style={darkgray176},
xmin=0.5, xmax=7.5,
xtick style={color=black},
xtick={1,2,3,4,5,6,7},
xticklabel style={rotate=75.0},
xticklabels={No Attack, Greedy Heuristic, Decomposed Heuristic, DDPG, Ablation | Low Level, Ablation | High Level, \textbf{\textit{HMARL}}},
y grid style={darkgray176},
ylabel={Total Travel Time},
ymajorgrids,
ymin=2930445.75, ymax=3146161.25,
yminorgrids,
ytick style={color=black}
]
\addplot [black]
table {%
0.75 2980403
1.25 2980403
1.25 2991944.25
0.75 2991944.25
0.75 2980403
};
\addplot [black]
table {%
1 2980403
1 2965320
};
\addplot [black]
table {%
1 2991944.25
1 3005866
};
\addplot [black]
table {%
0.875 2965320
1.125 2965320
};
\addplot [black]
table {%
0.875 3005866
1.125 3005866
};
\addplot [black]
table {%
1.75 2969359
2.25 2969359
2.25 2978304.75
1.75 2978304.75
1.75 2969359
};
\addplot [black]
table {%
2 2969359
2 2960940
};
\addplot [black]
table {%
2 2978304.75
2 2983394
};
\addplot [black]
table {%
1.875 2960940
2.125 2960940
};
\addplot [black]
table {%
1.875 2983394
2.125 2983394
};
\addplot [black]
table {%
2.75 2965428.75
3.25 2965428.75
3.25 2976668.75
2.75 2976668.75
2.75 2965428.75
};
\addplot [black]
table {%
3 2965428.75
3 2958714
};
\addplot [black]
table {%
3 2976668.75
3 2986716
};
\addplot [black]
table {%
2.875 2958714
3.125 2958714
};
\addplot [black]
table {%
2.875 2986716
3.125 2986716
};
\addplot [black]
table {%
3.75 2997666.75
4.25 2997666.75
4.25 3026279.75
3.75 3026279.75
3.75 2997666.75
};
\addplot [black]
table {%
4 2997666.75
4 2976721
};
\addplot [black]
table {%
4 3026279.75
4 3051419
};
\addplot [black]
table {%
3.875 2976721
4.125 2976721
};
\addplot [black]
table {%
3.875 3051419
4.125 3051419
};
\addplot [black, mark=o, mark size=3, mark options={solid,fill opacity=0}, only marks]
table {%
4 2940251
4 2941637
};
\addplot [black]
table {%
4.75 3009778.25
5.25 3009778.25
5.25 3023368.75
4.75 3023368.75
4.75 3009778.25
};
\addplot [black]
table {%
5 3009778.25
5 2995039
};
\addplot [black]
table {%
5 3023368.75
5 3036414
};
\addplot [black]
table {%
4.875 2995039
5.125 2995039
};
\addplot [black]
table {%
4.875 3036414
5.125 3036414
};
\addplot [black, mark=o, mark size=3, mark options={solid,fill opacity=0}, only marks]
table {%
5 3047143
};
\addplot [black]
table {%
5.75 2983808
6.25 2983808
6.25 3004979
5.75 3004979
5.75 2983808
};
\addplot [black]
table {%
6 2983808
6 2976072
};
\addplot [black]
table {%
6 3004979
6 3020840
};
\addplot [black]
table {%
5.875 2976072
6.125 2976072
};
\addplot [black]
table {%
5.875 3020840
6.125 3020840
};
\addplot [black]
table {%
6.75 3034240.75
7.25 3034240.75
7.25 3052667
6.75 3052667
6.75 3034240.75
};
\addplot [black]
table {%
7 3034240.75
7 3015705
};
\addplot [black]
table {%
7 3052667
7 3071361
};
\addplot [black]
table {%
6.875 3015705
7.125 3015705
};
\addplot [black]
table {%
6.875 3071361
7.125 3071361
};
\addplot [black, mark=o, mark size=3, mark options={solid,fill opacity=0}, only marks]
table {%
7 3084343
7 3081885
7 3097811
7 3136356
};
\addplot [darkorange25512714]
table {%
0.75 2988079
1.25 2988079
};
\addplot [darkorange25512714]
table {%
1.75 2973908.5
2.25 2973908.5
};
\addplot [darkorange25512714]
table {%
2.75 2972304
3.25 2972304
};
\addplot [darkorange25512714]
table {%
3.75 3012833.5
4.25 3012833.5
};
\addplot [darkorange25512714]
table {%
4.75 3017372.5
5.25 3017372.5
};
\addplot [darkorange25512714]
table {%
5.75 2993470.5
6.25 2993470.5
};
\addplot [darkorange25512714]
table {%
6.75 3040174.5
7.25 3040174.5
};
\end{axis}

\end{tikzpicture}}
        \caption{$B = 5$}
        \label{fig:sf-b5}
    \end{subfigure}
    \hfill
    \begin{subfigure}[b]{0.245\textwidth}
        \centering
        \resizebox{\linewidth}{!}{\begin{tikzpicture}

\definecolor{darkgray176}{RGB}{176,176,176}
\definecolor{darkorange25512714}{RGB}{255,127,14}

\begin{axis}[
tick align=outside,
tick pos=left,
x grid style={darkgray176},
xmin=0.5, xmax=7.5,
xtick style={color=black},
xtick={1,2,3,4,5,6,7},
xticklabel style={rotate=75.0},
xticklabels={No Attack, Greedy Heuristic, Decomposed Heuristic, DDPG, Ablation | Low Level, Ablation | High Level, \textbf{\textit{HMARL}}},
y grid style={darkgray176},
ymajorgrids,
ymin=2945922.05, ymax=3388230.95,
yminorgrids,
ytick style={color=black}
]
\addplot [black]
table {%
0.75 2980486.25
1.25 2980486.25
1.25 2991218
0.75 2991218
0.75 2980486.25
};
\addplot [black]
table {%
1 2980486.25
1 2968582
};
\addplot [black]
table {%
1 2991218
1 3001388
};
\addplot [black]
table {%
0.875 2968582
1.125 2968582
};
\addplot [black]
table {%
0.875 3001388
1.125 3001388
};
\addplot [black]
table {%
1.75 2977800.5
2.25 2977800.5
2.25 2987663.5
1.75 2987663.5
1.75 2977800.5
};
\addplot [black]
table {%
2 2977800.5
2 2969809
};
\addplot [black]
table {%
2 2987663.5
2 2997788
};
\addplot [black]
table {%
1.875 2969809
2.125 2969809
};
\addplot [black]
table {%
1.875 2997788
2.125 2997788
};
\addplot [black]
table {%
2.75 2975993
3.25 2975993
3.25 2986737.5
2.75 2986737.5
2.75 2975993
};
\addplot [black]
table {%
3 2975993
3 2966027
};
\addplot [black]
table {%
3 2986737.5
3 2996027
};
\addplot [black]
table {%
2.875 2966027
3.125 2966027
};
\addplot [black]
table {%
2.875 2996027
3.125 2996027
};
\addplot [black]
table {%
3.75 3019349.5
4.25 3019349.5
4.25 3077576.5
3.75 3077576.5
3.75 3019349.5
};
\addplot [black]
table {%
4 3019349.5
4 2993590
};
\addplot [black]
table {%
4 3077576.5
4 3126178
};
\addplot [black]
table {%
3.875 2993590
4.125 2993590
};
\addplot [black]
table {%
3.875 3126178
4.125 3126178
};
\addplot [black]
table {%
4.75 3171779.5
5.25 3171779.5
5.25 3196041.5
4.75 3196041.5
4.75 3171779.5
};
\addplot [black]
table {%
5 3171779.5
5 3153171
};
\addplot [black]
table {%
5 3196041.5
5 3216390
};
\addplot [black]
table {%
4.875 3153171
5.125 3153171
};
\addplot [black]
table {%
4.875 3216390
5.125 3216390
};
\addplot [black, mark=o, mark size=3, mark options={solid,fill opacity=0}, only marks]
table {%
5 3232655
};
\addplot [black]
table {%
5.75 2995223.25
6.25 2995223.25
6.25 3006205.25
5.75 3006205.25
5.75 2995223.25
};
\addplot [black]
table {%
6 2995223.25
6 2986151
};
\addplot [black]
table {%
6 3006205.25
6 3021942
};
\addplot [black]
table {%
5.875 2986151
6.125 2986151
};
\addplot [black]
table {%
5.875 3021942
6.125 3021942
};
\addplot [black, mark=o, mark size=3, mark options={solid,fill opacity=0}, only marks]
table {%
6 2975928
};
\addplot [black]
table {%
6.75 3211447.25
7.25 3211447.25
7.25 3235765.5
6.75 3235765.5
6.75 3211447.25
};
\addplot [black]
table {%
7 3211447.25
7 3184306
};
\addplot [black]
table {%
7 3235765.5
7 3264222
};
\addplot [black]
table {%
6.875 3184306
7.125 3184306
};
\addplot [black]
table {%
6.875 3264222
7.125 3264222
};
\addplot [black, mark=o, mark size=3, mark options={solid,fill opacity=0}, only marks]
table {%
7 3363605
7 3368126
7 3280375
7 3303229
7 3283716
7 3275664
7 3275792
};
\addplot [darkorange25512714]
table {%
0.75 2986156.5
1.25 2986156.5
};
\addplot [darkorange25512714]
table {%
1.75 2984460.5
2.25 2984460.5
};
\addplot [darkorange25512714]
table {%
2.75 2981828.5
3.25 2981828.5
};
\addplot [darkorange25512714]
table {%
3.75 3048120.5
4.25 3048120.5
};
\addplot [darkorange25512714]
table {%
4.75 3183649.5
5.25 3183649.5
};
\addplot [darkorange25512714]
table {%
5.75 3000898.5
6.25 3000898.5
};
\addplot [darkorange25512714]
table {%
6.75 3222612
7.25 3222612
};
\end{axis}

\end{tikzpicture}}
        \caption{$B = 10$}
    \end{subfigure}
    \hfill
    \begin{subfigure}[b]{0.245\textwidth}
        \centering
        \resizebox{\linewidth}{!}{\begin{tikzpicture}

\definecolor{darkgray176}{RGB}{176,176,176}
\definecolor{darkorange25512714}{RGB}{255,127,14}

\begin{axis}[
tick align=outside,
tick pos=left,
x grid style={darkgray176},
xmin=0.5, xmax=7.5,
xtick style={color=black},
xtick={1,2,3,4,5,6,7},
xticklabel style={rotate=75.0},
xticklabels={No Attack, Greedy Heuristic, Decomposed Heuristic, DDPG, Ablation | Low Level, Ablation | High Level, \textbf{\textit{HMARL}}},
y grid style={darkgray176},
ymajorgrids,
ymin=2933547, ymax=3742245,
yminorgrids,
ytick style={color=black}
]
\addplot [black]
table {%
0.75 2979675.75
1.25 2979675.75
1.25 2993115.25
0.75 2993115.25
0.75 2979675.75
};
\addplot [black]
table {%
1 2979675.75
1 2971197
};
\addplot [black]
table {%
1 2993115.25
1 3005806
};
\addplot [black]
table {%
0.875 2971197
1.125 2971197
};
\addplot [black]
table {%
0.875 3005806
1.125 3005806
};
\addplot [black]
table {%
1.75 2987399.5
2.25 2987399.5
2.25 2996704.5
1.75 2996704.5
1.75 2987399.5
};
\addplot [black]
table {%
2 2987399.5
2 2974291
};
\addplot [black]
table {%
2 2996704.5
2 3007634
};
\addplot [black]
table {%
1.875 2974291
2.125 2974291
};
\addplot [black]
table {%
1.875 3007634
2.125 3007634
};
\addplot [black, mark=o, mark size=3, mark options={solid,fill opacity=0}, only marks]
table {%
2 2970306
};
\addplot [black]
table {%
2.75 2982581.5
3.25 2982581.5
3.25 2994809.25
2.75 2994809.25
2.75 2982581.5
};
\addplot [black]
table {%
3 2982581.5
3 2973369
};
\addplot [black]
table {%
3 2994809.25
3 3011016
};
\addplot [black]
table {%
2.875 2973369
3.125 2973369
};
\addplot [black]
table {%
2.875 3011016
3.125 3011016
};
\addplot [black]
table {%
3.75 3127989
4.25 3127989
4.25 3192029.25
3.75 3192029.25
3.75 3127989
};
\addplot [black]
table {%
4 3127989
4 3096071
};
\addplot [black]
table {%
4 3192029.25
4 3268951
};
\addplot [black]
table {%
3.875 3096071
4.125 3096071
};
\addplot [black]
table {%
3.875 3268951
4.125 3268951
};
\addplot [black, mark=o, mark size=3, mark options={solid,fill opacity=0}, only marks]
table {%
4 3308298
4 3297587
};
\addplot [black]
table {%
4.75 3354146.25
5.25 3354146.25
5.25 3375239.5
4.75 3375239.5
4.75 3354146.25
};
\addplot [black]
table {%
5 3354146.25
5 3327629
};
\addplot [black]
table {%
5 3375239.5
5 3401896
};
\addplot [black]
table {%
4.875 3327629
5.125 3327629
};
\addplot [black]
table {%
4.875 3401896
5.125 3401896
};
\addplot [black]
table {%
5.75 3004331.5
6.25 3004331.5
6.25 3019782.25
5.75 3019782.25
5.75 3004331.5
};
\addplot [black]
table {%
6 3004331.5
6 2981838
};
\addplot [black]
table {%
6 3019782.25
6 3039029
};
\addplot [black]
table {%
5.875 2981838
6.125 2981838
};
\addplot [black]
table {%
5.875 3039029
6.125 3039029
};
\addplot [black]
table {%
6.75 3453898.25
7.25 3453898.25
7.25 3567408.25
6.75 3567408.25
6.75 3453898.25
};
\addplot [black]
table {%
7 3453898.25
7 3354109
};
\addplot [black]
table {%
7 3567408.25
7 3705486
};
\addplot [black]
table {%
6.875 3354109
7.125 3354109
};
\addplot [black]
table {%
6.875 3705486
7.125 3705486
};
\addplot [darkorange25512714]
table {%
0.75 2986585.5
1.25 2986585.5
};
\addplot [darkorange25512714]
table {%
1.75 2993283
2.25 2993283
};
\addplot [darkorange25512714]
table {%
2.75 2987421
3.25 2987421
};
\addplot [darkorange25512714]
table {%
3.75 3158305.5
4.25 3158305.5
};
\addplot [darkorange25512714]
table {%
4.75 3363000.5
5.25 3363000.5
};
\addplot [darkorange25512714]
table {%
5.75 3011272
6.25 3011272
};
\addplot [darkorange25512714]
table {%
6.75 3505886
7.25 3505886
};
\end{axis}

\end{tikzpicture}}
        \caption{$B = 15$}
    \end{subfigure}
    \hfill
    \begin{subfigure}[b]{0.245\textwidth}
        \centering
        \resizebox{\linewidth}{!}{\begin{tikzpicture}

\definecolor{darkgray176}{RGB}{176,176,176}
\definecolor{darkorange25512714}{RGB}{255,127,14}

\begin{axis}[
tick align=outside,
tick pos=left,
x grid style={darkgray176},
xmin=0.5, xmax=7.5,
xtick style={color=black},
xtick={1,2,3,4,5,6,7},
xticklabel style={rotate=75.0},
xticklabels={No Attack, Greedy Heuristic, Decomposed Heuristic, DDPG, Ablation | Low Level, Ablation | High Level, \textbf{\textit{HMARL}}},
y grid style={darkgray176},
ymajorgrids,
ymin=2853891.15, ymax=5417569.85,
yminorgrids,
ytick style={color=black}
]
\addplot [black]
table {%
0.75 2982567
1.25 2982567
1.25 2994666.5
0.75 2994666.5
0.75 2982567
};
\addplot [black]
table {%
1 2982567
1 2970422
};
\addplot [black]
table {%
1 2994666.5
1 3006299
};
\addplot [black]
table {%
0.875 2970422
1.125 2970422
};
\addplot [black]
table {%
0.875 3006299
1.125 3006299
};
\addplot [black]
table {%
1.75 3038554
2.25 3038554
2.25 3057483.75
1.75 3057483.75
1.75 3038554
};
\addplot [black]
table {%
2 3038554
2 3029130
};
\addplot [black]
table {%
2 3057483.75
2 3084956
};
\addplot [black]
table {%
1.875 3029130
2.125 3029130
};
\addplot [black]
table {%
1.875 3084956
2.125 3084956
};
\addplot [black, mark=o, mark size=3, mark options={solid,fill opacity=0}, only marks]
table {%
2 3092135
2 3088709
2 3093361
};
\addplot [black]
table {%
2.75 3055427
3.25 3055427
3.25 3064436
2.75 3064436
2.75 3055427
};
\addplot [black]
table {%
3 3055427
3 3042437
};
\addplot [black]
table {%
3 3064436
3 3074264
};
\addplot [black]
table {%
2.875 3042437
3.125 3042437
};
\addplot [black]
table {%
2.875 3074264
3.125 3074264
};
\addplot [black, mark=o, mark size=3, mark options={solid,fill opacity=0}, only marks]
table {%
3 3085729
3 3086045
};
\addplot [black]
table {%
3.75 3272059.25
4.25 3272059.25
4.25 3298746.25
3.75 3298746.25
3.75 3272059.25
};
\addplot [black]
table {%
4 3272059.25
4 3257646
};
\addplot [black]
table {%
4 3298746.25
4 3327255
};
\addplot [black]
table {%
3.875 3257646
4.125 3257646
};
\addplot [black]
table {%
3.875 3327255
4.125 3327255
};
\addplot [black]
table {%
4.75 3532359.5
5.25 3532359.5
5.25 3593297.25
4.75 3593297.25
4.75 3532359.5
};
\addplot [black]
table {%
5 3532359.5
5 3494223
};
\addplot [black]
table {%
5 3593297.25
5 3684664
};
\addplot [black]
table {%
4.875 3494223
5.125 3494223
};
\addplot [black]
table {%
4.875 3684664
5.125 3684664
};
\addplot [black, mark=o, mark size=3, mark options={solid,fill opacity=0}, only marks]
table {%
5 3750499
5 3778567
5 3692158
};
\addplot [black]
table {%
5.75 3138549.5
6.25 3138549.5
6.25 3210530.25
5.75 3210530.25
5.75 3138549.5
};
\addplot [black]
table {%
6 3138549.5
6 3085601
};
\addplot [black]
table {%
6 3210530.25
6 3236608
};
\addplot [black]
table {%
5.875 3085601
6.125 3085601
};
\addplot [black]
table {%
5.875 3236608
6.125 3236608
};
\addplot [black]
table {%
6.75 4389605.25
7.25 4389605.25
7.25 4639163.25
6.75 4639163.25
6.75 4389605.25
};
\addplot [black]
table {%
7 4389605.25
7 4106503
};
\addplot [black]
table {%
7 4639163.25
7 4875088
};
\addplot [black]
table {%
6.875 4106503
7.125 4106503
};
\addplot [black]
table {%
6.875 4875088
7.125 4875088
};
\addplot [black, mark=o, mark size=3, mark options={solid,fill opacity=0}, only marks]
table {%
7 5096323
7 5301039
};
\addplot [darkorange25512714]
table {%
0.75 2987809
1.25 2987809
};
\addplot [darkorange25512714]
table {%
1.75 3043419.5
2.25 3043419.5
};
\addplot [darkorange25512714]
table {%
2.75 3060523
3.25 3060523
};
\addplot [darkorange25512714]
table {%
3.75 3283913.5
4.25 3283913.5
};
\addplot [darkorange25512714]
table {%
4.75 3557800.5
5.25 3557800.5
};
\addplot [darkorange25512714]
table {%
5.75 3184553
6.25 3184553
};
\addplot [darkorange25512714]
table {%
6.75 4480706
7.25 4480706
};
\end{axis}

\end{tikzpicture}}
        \caption{$B = 30$}
    \end{subfigure}

\caption{Ablation study of HMARL on the \textbf{Sioux Falls} network. \emph{``No Attack''} pertains to no attack on the network. \emph{``Greedy Heuristic''} is a network greedy (see Section~\ref{sec:heuristics}) attack. \emph{``DDPG''} applies the general-purpose DDPG algorithm network-wide. In the remaining columns, the network is divided into \textbf{four} components. In \emph{``Decomposed Heuristic,''} the low-level actors are low-level greedy agents, with the high-level being a proportional allocation to the number of vehicles in each component. In \emph{``Ablation | Low Level,''} the high-level agent is the proportional allocation heuristic, while its low-level is the MADDPG approach. In \emph{``Ablation | High Level,''} the low-level is the greedy heuristic, while the high-level is a DDPG allocator RL agent. \textbf{\emph{``HMARL''}} is our HMARL approach. Here, the low-level MADDPG and high-level DDPG components have been trained simultaneously.}
\label{fig:siouxfalls}

\end{figure*}
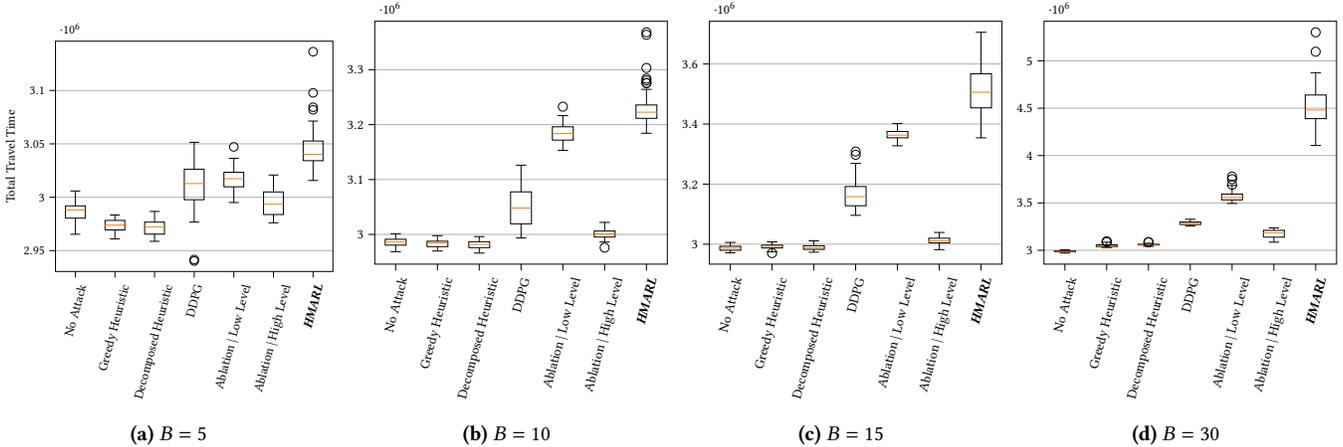

\begin{table}[t]
    \centering
    \caption{Neural Network Architecture}
    \label{tab:nn_arch}
        \begin{tabular}{|l|c|}
            \hline
            Hyperparameter & Value \\
            \hline
            \rowcolor{Gray} \multicolumn{2}{|c|}{High-Level} \\
            \hline
            \rowcolor{LightGray} \multicolumn{2}{|c|}{Actor $\mu_H$} \\
            Number of hidden layers & 2 \\
            Sizes of layers & [256, 128] \\
            Activation function & ReLU \\
            Optimizer & Adam \\
            \rowcolor{LightGray} \multicolumn{2}{|c|}{Critic $Q_H$} \\
            Number of hidden layers & 2 \\
            Sizes of hidden layers & [128, 128] \\
            Activation function & ReLU \\
            Optimizer & Adam \\
            \hline
            \rowcolor{Gray} \multicolumn{2}{|c|}{Low-Level} \\
            \hline
            \rowcolor{LightGray} \multicolumn{2}{|c|}{Actor $\mu_k$} \\
            Number of hidden layers & 2 \\
            Sizes of hidden layers & [512, 512] \\
            Activation function & [ReLU, ReLU, Sigmoid] \\
            Optimizer & Adam \\
            \rowcolor{LightGray} \multicolumn{2}{|c|}{Critic $Q_k$} \\
            Number of hidden layers & 2 \\
            Sizes of hidden layers & [128, 128] \\
            Activation function & ReLU \\
            Optimizer & Adam \\
            \hline
        \end{tabular}
\end{table}

After the initialization of the environment, as the HMARL is off-policy, it can draw experiences of states, actions, next states, and rewards from the environment by taking either random actions or by taking Ornstein-Uhlenbeck~\cite{ou} noise added to actions outputted by the low-level agent. Using these experiences, all actors and critics can be updated simultaneously. Algorithm~\ref{alg:hmarl} shows the training workflow of the HMARL.

When agents are trained simultaneously, the low-level agent should have lower learning rates as it needs the high-level agent to learn its behavior but should account for more steps in the future with a higher discount factor $\gamma$.

Figure~\ref{fig:siouxfalls} shows the result of the training with an ablation study on the Sioux Falls, ND transportation network \cite{transportationnetworks}. This network has 24 nodes and 76 edge links. We ran HMARL with different attack budgets. As expected, the HMARL performs better by 10-50\% depending on the budget, making it a viable solution to the scalability of Deep Reinforcement Algorithms.

\begin{algorithm}
\caption{Hierarchical Multi-Agent Reinforcement Learning}
\label{alg:hmarl}

\begin{algorithmic}
\Require A road network graph $G=(V, E)$; Set of Riders $R$;

\State Initialize environment $env$
\State Initialize Replay Buffer $E$.
\State Run K-Means Clustering to acquire components

\State $s \gets env.reset()$
\For{step $\Rightarrow$ total steps}
    \State $\vect{\hat{b}} \gets \mu_H(\hat{s})$
    \State $\vect{a} \gets \odot\mu_k(\hat{b}, s) + \mathcal{N}$
    \State Normalize $\vect{a}$

    \State $s', r \gets env.step(\vect{a})$
    \State $E \gets E \cup \langle s, s', \vect{a}, r \rangle$
    \State $s \gets s'$
    \If{$env.done()$}
        \State $s \gets env.reset()$
    \EndIf

    \State Sample $\hat{E} \sim E$
    \State Update $Q_H$, $\mu_H$, $Q_k, \mu_k \forall_{k \in K}$ with $\hat{E}$
    
\EndFor
\end{algorithmic}
\end{algorithm}

\section{Discussion}
\label{sec:discussion}

First, we conducted a hyperparameter optimization with a simple grid search and reported only the hyperparameters that work best. Further evaluation of hyperparameters with more sophisticated search mechanisms, such as Bayesian Search~\cite{bayesian}, is required. 

Based on the experiments conducted, it is important to note that the shortest path routing approach for traffic does not always result in an optimal network flow solution due to network congestion. It is possible for a non-optimal attack to actually reduce the total travel time of the vehicles by decreasing congestion on regular congested paths. For example, see Figure~\ref{fig:sf-b5} with no attack (``No Attack'') and Greedy (``Greedy Heuristic'', ``Decomposed Heuristic'') strategies.

Further, the starting nodes for the k-means clustering can impact the training process and need to be thoroughly examined. There should be a correlation between certain metrics, such as the maximum diameter of components, the number of nodes in components, the balance of nodes and edges between the components of the outputted clusters, and the performance of HMARL, which has not yet been analyzed. Additionally, other graph decomposition methods have not been evaluated. It is worth considering a different decomposition algorithm that tends to produce more balanced components, as this may lead to a more stable training process and better performance of the HMARL.

There is a trade-off between the number of components, the performance of high-level, low-level, and the hierarchical approach. With too many components ($|K| \rightarrow |V|$), or with too few components ($|K| \rightarrow 1$), the hierarchical approach will be equivalent to a single-agent RL. The best number of components should be extracted experimentally.

When analyzing heuristics and HMARL, it is essential to consider the sparsity of vehicles. The rider data in Sioux Falls is dense, with 100-500 vehicles traveling between each pair of nodes. On the contrary, the rider data in Eastern Massachusetts is sparse, with no vehicles present for over 75\% of pairs of nodes.

Currently, the feature extractor of the state representation is a Multi-Layer Perceptron. As the state is inherently a graph, Graph Convolutional Networks~\cite{morris2021weisfeiler} or Graph Attention Networks~\cite{velickovic2018graph} can be incorporated to improve the accuracy of the $Q$ and $\mu$ functions.

\subsection{Conclusion}

Our research focused on the impact of adversarial influence on transportation networks. We investigated how drivers could be manipulated by injecting false information into navigation apps through a computational approach. We developed an adversarial model that included a threat actor capable of manipulating drivers by increasing their perceived travel times, leading them to take suboptimal and longer routes. To accomplish this, we created a computational framework based on Hierarchical Multi-Agent Deep Reinforcement Learning (HMARL) to determine an optimal strategy for data manipulation. Our simulation of the Sioux Falls, ND transportation network showed that the adversary could increase the total travel time of all drivers by 50\%.

\subsection{Future Direction}
To effectively combat data attacks, it is crucial to have a robust defense mechanism. One of the most significant hurdles is precisely identifying data manipulation attacks. Our research has yielded positive outcomes, and we aim to tackle the problem of detecting such false-data injection attacks by utilizing sophisticated machine-learning methods and competitive multi-agent reinforcement learning algorithms.

\begin{acks}
This material is based upon work sponsored by the National Science Foundation under Grants CNS-1952011, IIS-1905558, IIS-1903207, and IIS-2214141 and by the Army Research Office under Grant W911NF1810208. 
Any opinions, findings, and conclusions or recommendations expressed in this material are those of the authors and do not necessarily reflect the views of the National Science Foundation and of the Army Research Office.
\end{acks}

\balance

\bibliographystyle{ACM-Reference-Format} 
\bibliography{main}

\end{document}